\documentclass[11pt]{article}
\usepackage[letterpaper,margin=0.86in]{geometry}
\usepackage{amsmath,amssymb,amsthm,mathtools,bm}
\usepackage{microtype}
\usepackage{enumitem}
\usepackage{xspace}
\usepackage{xcolor}
\usepackage{hyperref}
\usepackage{cleveref}
\usepackage{natbib}
\usepackage{authblk}

\hypersetup{colorlinks=true,linkcolor=blue!40!black,citecolor=blue!40!black,urlcolor=blue!40!black}
\setlist{leftmargin=*,itemsep=0.15em,topsep=0.25em}

\newcommand{\ELVAE}{\textsc{ELVAE}\xspace}
\newcommand{\E}{\mathbb{E}}
\newcommand{\Var}{\operatorname{Var}}
\newcommand{\KL}{D_{\mathrm{KL}}}
\newcommand{\NIG}{\operatorname{NIG}}
\newcommand{\IG}{\operatorname{InvGamma}}
\newcommand{\Normal}{\mathcal{N}}
\newcommand{\can}{\mathrm{can}}
\newcommand{\uepi}{u_{\mathrm{epi}}}
\newcommand{\uvar}{u_{\mathrm{var}}}
\newcommand{\Rcan}{\mathcal{R}_{\mathrm{can}}}
\newcommand{\Lrec}{\mathcal{L}_{\mathrm{rec}}}

\newtheorem{theorem}{Theorem}
\newtheorem{proposition}{Proposition}
\newtheorem{corollary}{Corollary}

\theoremstyle{definition}

\theoremstyle{remark}
\newtheorem{remark}{Remark}

\title{\textbf{Theoretical Study on the Evidential Learning-based Variational Autoencoder}\\}
\author{Ge Wang}
\affil{Rensselaer Polytechnic Institute, Troy, New York, USA}
\date{August 18, 2026}

\begin{document}
\maketitle

\begin{abstract}
A normal--inverse-gamma (NIG) latent hierarchy has four parameters, but its induced latent law does not identify all four hierarchy parameters. For \(\sigma^2 \sim \mathrm{InvGamma}(\alpha,\beta)\), \(\mu \mid \sigma^2 \sim \mathcal{N}(\gamma,\sigma^2/\nu)\), and \(z \mid \mu,\sigma^2 \sim \mathcal{N}(\mu,\sigma^2)\), the marginal distribution of \(z\) depends on \((\nu,\beta)\) only through \(c=\beta(1+1/\nu)\). Thus, the reconstruction-visible parameter space is the three-dimensional quotient with coordinates \((\gamma,\alpha,c)\), while \((\nu,\beta)\) contain a one-dimensional fiber degree of freedom. This marginal degeneracy is related to the known NIG degeneracy in deep evidential regression. The present contribution is the quotient-and-section formulation for a latent bottleneck and its variational reduction.

Given a fixed variational objective, the representative on each fiber is not an additional free modeling choice: exact partial minimization of the forward KL divergence to a complete hierarchical prior \(p_0=\mathrm{NIG}(\gamma_0,\nu_0,\alpha_0,\beta_0)\), with \(c_0=\beta_0(1+1/\nu_0)\), selects the unique prior-relative representative
\(\nu_{\mathrm{can}}=B-\alpha_0-\tfrac12+\sqrt{(B-\alpha_0-\tfrac12)^2+2B}\), where \(B=\tfrac{\nu_0}{2}+\tfrac{\alpha}{c}\left[\beta_0+\tfrac{\nu_0}{2}(\gamma-\gamma_0)^2\right]\), and \(\beta_{\mathrm{can}}=c\nu_{\mathrm{can}}/(1+\nu_{\mathrm{can}})\).

Letting \(\rho_0=2\beta_0/\nu_0\) and \(T=c/\{\alpha[(\gamma-\gamma_0)^2+\rho_0]\}\), we show that \(1/\nu_{\mathrm{can}}=g_{\nu_0,\alpha_0}(T)\) is an explicit strictly increasing function. For \(\alpha>1\), the reported ratio \(u_{\mathrm{epi}}/u_{\mathrm{var}}=1/\nu_{\mathrm{can}}\) is completely determined by the quotient state and the specified prior; for rank-only use, \(T\) contains exactly the same coordinatewise ordinal information under a common prior calibration. The prior gauge is compressed rather than eliminated. Ordinal dependence is governed by \((\gamma_0,\rho_0)\), whereas \((\nu_0,\alpha_0)\) determine numerical calibration and the analytic ceiling of \(1/\nu_{\mathrm{can}}\). Along a fixed prior marginal law, the ordinal parameter satisfies \(\rho_0\in(0,2c_0)\). Detailed derivations are deferred to the appendices.
\end{abstract}


\section{Introduction}
\label{sec:introduction}

Variational autoencoders replace a deterministic latent representation by an amortized probability law and optimize a reconstruction term together with a distributional regularizer \citep{kingma2014,rezende2014}. Evidential learning takes a related higher-order step: a network predicts parameters of a probability distribution over lower-order distributional parameters. In regression, the normal--inverse-gamma (NIG) family is convenient because it is conjugate to an unknown Gaussian mean and variance and supports analytic uncertainty decompositions \citep{amini2020}. Recent analyses nevertheless emphasize that assigning a probabilistic interpretation to a network output does not by itself make that quantity identifiable from the loss \citep{bengs2022,meinert2023}.

The issue studied here is structural. Does the four-parameter posterior of the NIG hierarchy provide four independently identifiable latent coordinates? It does not. After marginalizing the hierarchy, one degree of freedom disappears exactly. An entire one-dimensional family of NIG quadruples induces the same latent distribution and hence the same reconstruction functional.

This marginal degeneracy is not, by itself, new to evidential modeling. In deep evidential regression (DER), the Student--\(t\) marginal likelihood likewise depends on the NIG parameters \((\nu,\beta)\) only through the compensating combination \(\beta(1+\nu)/\nu\) \citep{amini2020}. Subsequent analyses have highlighted the associated overparameterization and loss-induced behavior \citep{meinert2023}. In the original DER objective, the likelihood degeneracy is supplemented by a residual-weighted evidence penalty, \(\lvert y-\gamma\rvert(2\nu+\alpha)\) \citep{amini2020}. The contribution here is different: we formulate the latent NIG hierarchy explicitly as a quotient-and-section problem and show that, for a hierarchical variational objective, the fiber can be eliminated by exact partial minimization of the forward KL divergence to a fixed NIG prior.

In other words, we treat the hierarchy as a fibered representation over a three-dimensional quotient. The reconstruction-visible coordinates are \((\gamma,\alpha,c)\), where
\begin{equation}
 c=\beta\left(1+\frac1\nu\right).
 \label{eq:cdef_main}
\end{equation}
The remaining coordinate changes only the hierarchical decomposition of a fixed marginal state. While the quotient reduction is intrinsic to the marginal likelihood geometry, the choice of a representative on each fiber is not.

A quotient space has no preferred section without additional structure. Here that structure is a fully specified hierarchical prior together with the forward KL divergence on the full NIG hierarchy already present in the variational objective. Given that objective,  the same forward KL can be minimized along each fiber. We call the resulting representative \emph{canonical relative to the chosen hierarchical prior and variational objective}.

The main results are as follows: (1) the four-parameter NIG bottleneck reduces exactly to the three quotient coordinates \((\gamma,\alpha,c)\) at the level of the marginal latent law; (2) a complete NIG prior and the forward KL term of the fixed variational objective define a unique prior-relative section of every fiber; (3) partial minimization over the fiber is exact at the level of the underlying variational family and preserves the optimum of the four-coordinate objective for every positive KL weight; (4) on the selected section, inverse allocation is an explicit deterministic transform of the quotient-geometric score \(T\); and
(5) the residual prior gauge is characterized rather than hidden: coordinatewise ordinal content depends on \((\gamma_0,\rho_0)\), numerical calibration on \((\nu_0,\alpha_0)\), and moving a prior along a fixed marginal-law fiber can still change the ordering.

The stationary allocation formula is related to the special-case calculation used diagnostically in the original \ELVAE formulation \citep{wang2026elvae}. The present contribution changes its role: the stationary point is generalized to a fixed NIG prior and promoted from a diagnostic quantity to the defining prior-relative section of the reduced variational family.
All detailed algebra is placed in the appendices. 

\section{NIG Latent Bottleneck as a Quotient Model} \label{sec:quotient} Consider one latent coordinate and suppress the coordinate index. For an input \(\bm y\), let \begin{align} \sigma^2\mid\bm y &\sim \IG(\alpha,\beta),\nonumber\\ \mu\mid\sigma^2,\bm y &\sim \Normal\!\left(\gamma,\frac{\sigma^2}{\nu}\right),\nonumber\\ z\mid\mu,\sigma^2 &\sim \Normal(\mu,\sigma^2), \label{eq:nig_hierarchy_main} \end{align} with \(\nu>0\), \(\alpha>0\), and \(\beta>0\). The NIG hierarchy is imposed only at the latent bottleneck; the observation model remains the decoder \(p_\theta(\bm y\mid\bm z)\). This separates the hierarchical uncertainty model from the observation-space likelihood and makes the identifiability question studied below a property of the latent representation. The quotient reduction below requires only \(\alpha>0\). Moment-based language is used later only under its stated moment conditions; in particular, the latent variance decomposition requires \(\alpha>1\). Let \(t_\kappa(m,s^2)\) denote a Student--\(t\) law with \(\kappa\) degrees of freedom, location \(m\), and \emph{squared scale} \(s^2\). Thus, its variance, when \(\kappa>2\), is \(s^2\kappa/(\kappa-2)\). \begin{proposition}[Reconstruction-visible coordinates] \label{prop:quotient} The marginal law of \(z\) induced by \cref{eq:nig_hierarchy_main} is \begin{equation} z\mid\bm y\sim t_{2\alpha}\!\left(\gamma,\frac{c}{\alpha}\right), \qquad c=\beta\left(1+\frac1\nu\right). \label{eq:student_main} \end{equation} Consequently, every downstream functional that depends on the hierarchy only through the marginal law of \(z\) depends on the NIG quadruple \((\gamma,\nu,\alpha,\beta)\) only through \((\gamma,\alpha,c)\). \end{proposition} The proof is given in \cref{app:collapse}. The same \((\nu,\beta)\) degeneracy occurs in the observation-space NIG marginal used by DER \citep{amini2020,meinert2023}. Proposition~\ref{prop:quotient} is not claimed as a new marginalization identity. Its role here is to expose the base space on which the subsequent quotient-and-section reduction is built. Proposition~\ref{prop:quotient} induces the equivalence relation \begin{equation} (\gamma,\nu,\alpha,\beta)\sim(\gamma',\nu',\alpha',\beta') \end{equation} whenever \begin{equation} \gamma=\gamma',\qquad \alpha=\alpha',\qquad \beta\left(1+\frac1\nu\right)= \beta'\left(1+\frac1{\nu'}\right). \label{eq:equiv} \end{equation} Each equivalence class is a one-dimensional fiber over \((\gamma,\alpha,c)\). A convenient parameterization is \begin{equation} \beta(\nu)=c\frac{\nu}{1+\nu},\qquad \nu\in(0,\infty), \label{eq:fiber_main} \end{equation} which is equivalently \(\beta\in(0,c)\) with \(\nu=\beta/(c-\beta)\). The quotient formulation therefore requires only three input-dependent coordinates, \begin{equation} \gamma(\bm y),\qquad \alpha(\bm y)>0,\qquad c(\bm y)>0. \label{eq:three_main} \end{equation} A fourth coordinate moves only along a fiber and cannot alter the exact marginal latent law while the base state is fixed. Although \(\alpha\) is a genuine coordinate of the exact marginal family, its tail-shape contribution may be weakly identified in near-Gaussian regimes, particularly along parameter variations that approximately preserve the marginal variance. This practical weak identification is distinct from the exact one-dimensional fiber degeneracy in \((\nu,\beta)\).
\begin{remark}[Fiber-invariant pathwise representation] \label{rem:pathwise} The quotient admits the equivalent collapsed hierarchy
\begin{equation} \omega^2\sim\IG(\alpha,c),\qquad z\mid\omega^2\sim\Normal(\gamma,\omega^2), \label{eq:collapsed_main} \end{equation} which contains no fiber coordinate. Hence, the common marginal law of every point on a fiber admits a pathwise realization using only the quotient coordinates \((\gamma,\alpha,c)\), rather than becoming fiber-invariant only after taking expectations. The derivation is given in \cref{app:collapse}. \end{remark}
\section{Prior-Relative Canonical Variational Reduction}
\label{sec:canonical}

Let the complete hierarchical reference distribution be
\begin{equation}
p_0(\mu,\sigma^2)=\NIG(\gamma_0,\nu_0,\alpha_0,\beta_0),
\qquad \nu_0,\alpha_0,\beta_0>0.
\label{eq:prior_main}
\end{equation}
For a fixed quotient state \((\gamma,\alpha,c)\), define its prior-relative canonical representative by
\begin{equation}
(\nu_{\can},\beta_{\can})
=
\arg\min_{\nu>0,\;\beta=c\nu/(1+\nu)}
\KL\!\left[
\NIG(\gamma,\nu,\alpha,\beta)\,\|\,p_0
\right].
\label{eq:selector_def}
\end{equation}

\begin{theorem}[Unique prior-relative canonical allocation]
\label{thm:canonical}
For every admissible \((\gamma,\alpha,c)\), define
\begin{equation}
B=
\frac{\nu_0}{2}
+\frac{\alpha}{c}
\left[
\beta_0+\frac{\nu_0}{2}(\gamma-\gamma_0)^2
\right].
\label{eq:B_main}
\end{equation}
Then, \cref{eq:selector_def} has the unique solution
\begin{align}
\nu_{\can}
&=B-\alpha_0-\frac12
+\sqrt{\left(B-\alpha_0-\frac12\right)^2+2B},
\label{eq:nucan_main}\\
\beta_{\can}
&=c\frac{\nu_{\can}}{1+\nu_{\can}}.
\label{eq:betacan_main}
\end{align}
The minimizer is global, interior, and real analytic as a function of \((\gamma,\alpha,c)\) on \(\mathbb R\times(0,\infty)\times(0,\infty)\).
\end{theorem}

The full divergence reduction and proof are deferred to \cref{app:kl,app:canonical}. The theorem supplies a unique section only after the full hierarchical prior and the forward KL geometry have been fixed.

\begin{remark}[Variational geometry, regularization level, and the residual global gauge]
\label{rem:prior_fiber}
Three structural specifications should be distinguished. First, once the variational objective \cref{eq:J4} is fixed, the forward divergence \(\KL(q\|p_0)\) is not an additional free convention: its fiber minimizer is exactly the representative produced by partial minimization of that objective. Another divergence can define another section, but if that section differs from \cref{eq:selector_def}, it is strictly suboptimal for the same forward-KL objective at fixed quotient coordinates.

Second, the section-selection problem presupposes that regularization is imposed on the full NIG hierarchy. Let \(c_0=\beta_0(1+1/\nu_0)\). If one first passes to the quotient and instead defines a different objective by regularizing the collapsed coordinate \(\IG(\alpha,c)\) directly against \(\IG(\alpha_0,c_0)\), the fiber coordinate is absent and no section remains to be selected. Such quotient-level regularization defines a different variational model; it is not an alternative section of \cref{eq:J4} and, in general, is not equal to the reduced hierarchical penalty \(\Rcan\).

Third, fixing only the prior marginal law \((\gamma_0,\alpha_0,c_0)\) does not determine \cref{eq:selector_def}: different choices of the prior fiber position \(\nu_0\) generally induce different sections. Thus ``canonical'' means canonical relative to the complete four-parameter prior, not relative to its marginal equivalence class. The reduction replaces an input-dependent fiber function by a globally specified reference gauge; it does not make that global choice intrinsic to the quotient. Section~\ref{sec:consequences} quantifies the residual dependence.
\end{remark}

The selector is variationally optimal in a stronger sense than uniqueness. Let \(\Lrec(\gamma,\alpha,c;\bm y)\) be any exact reconstruction loss determined only by the marginal law of \(z\), and consider
\begin{equation}
\mathcal J_4
=
\Lrec(\gamma,\alpha,c;\bm y)
+\lambda\,
\KL\!\left[
\NIG(\gamma,\nu,\alpha,\beta)\,\|\,p_0
\right],
\qquad \lambda>0,
\label{eq:J4}
\end{equation}
subject to \(c=\beta(1+1/\nu)\). Define
\begin{equation}
\Rcan(\gamma,\alpha,c)
=
\min_{\nu>0,\;\beta=c\nu/(1+\nu)}
\KL\!\left[
\NIG(\gamma,\nu,\alpha,\beta)\,\|\,p_0
\right]
\label{eq:Rcan_main}
\end{equation}
and
\begin{equation}
\mathcal J_3
=
\Lrec(\gamma,\alpha,c;\bm y)
+\lambda\Rcan(\gamma,\alpha,c).
\label{eq:J3}
\end{equation}
An explicit closed form for \(\Rcan\) is given in \cref{eq:app_Rcan_closed}.

\begin{theorem}[No-loss reduction of the variational family]
\label{thm:reduction}
For every \(\lambda>0\), partial minimization of the full NIG family over its fiber coordinate is exact:
\begin{equation}
\inf_{\gamma,\alpha,c,\nu}\mathcal J_4
=
\inf_{\gamma,\alpha,c}\mathcal J_3.
\label{eq:no_loss}
\end{equation}
For every fixed \((\gamma,\alpha,c)\), any noncanonical point on the same fiber has the same reconstruction term and strictly larger objective value.
\end{theorem}

Theorem~\ref{thm:reduction} is a statement about the underlying variational family after exact partial minimization. It does not assert identical optimization behavior or approximation capacity for arbitrary finite amortized neural-network parameterizations of the three- and four-output maps.

When \(\Lrec\) is the negative expected log likelihood and \(\lambda=1\), \cref{eq:no_loss} is exactly a negative-ELBO reduction under the hierarchical factorization
\begin{equation}
q(\mu,\sigma^2,z\mid\bm y)
=q(\mu,\sigma^2\mid\bm y)\,q(z\mid\mu,\sigma^2,\bm y),
\end{equation}
with
\begin{equation}
q(z\mid\mu,\sigma^2,\bm y)
=p(z\mid\mu,\sigma^2)=\Normal(\mu,\sigma^2).
\label{eq:conditional_match_main}
\end{equation}
The conditional \(z\)-level KL term therefore vanishes identically, leaving the expected negative log likelihood plus \(\KL[q(\mu,\sigma^2\mid\bm y)\|p_0]\). This is the sense in which the forward KL used to select the section is specified by the variational objective itself.

\begin{remark}[Two immediate consequences]
\label{rem:immediate}
For fixed \((\gamma,\alpha,c)\), multiplication of the fiber regularizer by any \(\lambda>0\) does not change its minimizer, so \(\nu_{\can}\) is conditionally weight-invariant; the outer optimum of \((\gamma,\alpha,c)\) may still depend on \(\lambda\). In addition, with \(\theta=(\gamma,\alpha,c)\), the interior stationarity condition gives the exact envelope identity
\begin{equation}
\nabla_\theta \Rcan(\theta)
=
\left.
\partial_\theta
\KL\!\left[
\NIG(\theta,\nu,\beta(\theta,\nu))\,\|\,p_0
\right]
\right|_{\nu=\nu_{\can}(\theta)},
\label{eq:envelope_main}
\end{equation}
where the partial derivative holds \(\nu\) fixed.
\end{remark}

The proof of Theorem~\ref{thm:reduction} and the two immediate consequences are collected in \cref{app:reduction}.

\section{Structural Consequences}
\label{sec:consequences}

The reduction separates three layers: the quotient coordinates identified by the marginal latent law, the prior-relative section selected by hierarchical divergence, and the second-moment interpretation available when that moment exists.

\subsection{Variance budget and internal allocation}
The quotient reduction itself requires only \(\alpha>0\). For the following variance decomposition we additionally assume \(\alpha>1\), so that \(\E[\sigma^2]\) is finite. For the prior-relative canonical representative,
\begin{align}
\uvar
&=\E[\sigma^2\mid\bm y]
=\frac{c}{\alpha-1}\frac{\nu_{\can}}{1+\nu_{\can}},
\label{eq:uvar_main}\\
\uepi
&=\Var(\mu\mid\bm y)
=\frac{c}{\alpha-1}\frac{1}{1+\nu_{\can}}.
\label{eq:uepi_main}
\end{align}
Thus
\begin{equation}
\boxed{
\uvar+\uepi=\frac{c}{\alpha-1},
\qquad
\frac{\uepi}{\uvar}=\frac1{\nu_{\can}}.
}
\label{eq:decomp_main}
\end{equation}
The quotient coordinate \(c/(\alpha-1)\) fixes the total latent variance. The prior-relative section fixes only its internal allocation.

\subsection{Self-consistency and dependence on the quotient state}
Let
\begin{equation}
c_0=\beta_0\left(1+\frac1{\nu_0}\right).
\end{equation}
At the prior base state \((\gamma,\alpha,c)=(\gamma_0,\alpha_0,c_0)\), the canonical representative is exactly the prior:
\begin{equation}
\nu_{\can}=\nu_0,
\qquad
\beta_{\can}=\beta_0.
\label{eq:self_main}
\end{equation}
Away from that state, \(\nu_{\can}\) remains input-dependent through the quotient coordinates, although it is no longer an independent amortized output.

\subsection{Ordinal geometry, transfer function, and residual gauge}
Define the prior radius floor
\begin{equation}
\rho_0:=\frac{2\beta_0}{\nu_0}>0
\label{eq:rho0_main}
\end{equation}
and the quotient-geometric coordinate
\begin{equation}
T
=
\frac{c}
{\alpha\left[(\gamma-\gamma_0)^2+\rho_0\right]}.
\label{eq:T_main}
\end{equation}
Then,
\begin{equation}
B(T)=\frac{\nu_0}{2}\left(1+\frac1T\right).
\label{eq:B_T_main}
\end{equation}
Let \(a=\alpha_0+1/2\). Rationalizing the reciprocal of \cref{eq:nucan_main} gives the explicit transfer function
\begin{equation}
\boxed{
\frac1{\nu_{\can}}
=g_{\nu_0,\alpha_0}(T)
:=
\frac{
\sqrt{(B(T)-a)^2+2B(T)}-[B(T)-a]
}{2B(T)}.
}
\label{eq:transfer_main}
\end{equation}
\footnote{For stable numerical evaluation, let \(d=B-a\) and \(D=d^2+2B\). Evaluate \(\nu_{\can}=d+\sqrt D\) when \(d\ge0\), and \(\nu_{\can}=2B/(\sqrt D-d)\) when \(d<0\); then invert if \(1/\nu_{\can}\) is required. This avoids subtractive cancellation in either regime.}

\begin{proposition}[Geometric re-encoding]
\label{prop:geometry}
The selector \(\nu_{\can}\) is strictly increasing in \(B\). The map \(T\mapsto B(T)\) is strictly decreasing, and \(B\mapsto1/\nu_{\can}(B)\) is strictly decreasing. Hence, their composition \(g_{\nu_0,\alpha_0}(T)\) is strictly increasing on \((0,\infty)\).
\end{proposition}

\begin{corollary}[No independent allocation information on the selected section]
\label{cor:noextra}
For a fixed complete prior, \(1/\nu_{\can}\) is a deterministic function of the quotient state \((\gamma,\alpha,c)\), equivalently of \(T\) together with the calibration pair \((\nu_0,\alpha_0)\). Thus, inverse canonical allocation is not an additional information channel beyond the quotient coordinates and prior. If \(\alpha>1\), both the ratio
\begin{equation}
\frac{\uepi}{\uvar}=g_{\nu_0,\alpha_0}(T)
\end{equation}
and the epistemic fraction
\begin{equation}
\frac{\uepi}{\uepi+\uvar}
=\frac{g_{\nu_0,\alpha_0}(T)}{1+g_{\nu_0,\alpha_0}(T)}
\end{equation}
are therefore determined by the same three quotient coordinates and the prior.

For any rank-only comparison to which a common transfer function applies, evaluating \(\nu_{\can}\) is unnecessary: ranking by \(1/\nu_{\can}\) is equivalent to ranking by the two-line score \(T\) in \cref{eq:T_main}.
\end{corollary}

The word \emph{coordinatewise} requires care. For a fixed latent coordinate compared across inputs or quotient states, the prior for that coordinate is fixed and Proposition~\ref{prop:geometry} gives exact ordinal equivalence. For ranking different latent dimensions \(k\) at one input, the same conclusion requires a common calibration map across the compared dimensions---for example, a shared coordinatewise NIG prior with common \((\nu_0,\alpha_0)\). If different dimensions use different \((\nu_{0,k},\alpha_{0,k})\), then distinct increasing transfer functions \(g_k\) can change cross-dimensional ranks even though each dimension separately remains monotone in its own \(T_k\).

\begin{corollary}[Ordinal dependence on the prior]
\label{cor:ordinal}
Under the equivalent prior parameterization
\begin{equation}
(\gamma_0,\nu_0,\alpha_0,\beta_0)
\longleftrightarrow
(\gamma_0,\rho_0,\nu_0,\alpha_0),
\qquad
\beta_0=\frac{\rho_0\nu_0}{2},
\label{eq:prior_reparam_main}
\end{equation}
the ordering induced by \(T\), and hence the ordering by \(1/\nu_{\can}\) under a common transfer function, depends on the prior through \((\gamma_0,\rho_0)\). In particular, any two complete priors with the same \((\gamma_0,\rho_0)\) induce the same ordering over any collection of quotient states to which each prior applies, although the numerical values of \(1/\nu_{\can}\) can differ through \((\nu_0,\alpha_0)\).
\end{corollary}

Corollary~\ref{cor:ordinal} does not eliminate the residual prior-fiber gauge in Remark~\ref{rem:prior_fiber}. Holding the prior marginal law \((\gamma_0,\alpha_0,c_0)\) fixed while moving its fiber position gives
\begin{equation}
\rho_0=\frac{2c_0}{1+\nu_0},
\qquad
0<\rho_0<2c_0,
\label{eq:rho0_fixed_marginal_main}
\end{equation}
so \(\rho_0\) changes with \(\nu_0\), and the ordering can change. The range in \cref{eq:rho0_fixed_marginal_main} is one-sided: as \(\nu_0\to\infty\), \(\rho_0\downarrow0\); as \(\nu_0\downarrow0\), \(\rho_0\uparrow2c_0\), but the endpoint is not attained. Thus a residual gauge motion that preserves the prior marginal Student--\(t\) law can approach the inverse-normalized-radius endpoint described below, but it cannot reach the large-\(\rho_0\) marginal-scale endpoint without changing that prior marginal law. The residual gauge is compressed, not neutralized.

For the common centered choice \(\gamma_0=0\), the ordinal prior dependence is reduced to the single positive scalar \(\rho_0\).

\begin{remark}[The radius floor as an interpolation parameter]
\label{rem:rho_interp}
The scalar \(\rho_0\) has a direct geometric role. For \(\gamma\neq\gamma_0\),
\begin{equation}
T\xrightarrow[\rho_0\downarrow0]{}
\frac{c}{\alpha(\gamma-\gamma_0)^2},
\label{eq:T_small_rho_main}
\end{equation}
a quotient-level inverse normalized-radius score, with \(\rho_0>0\) regularizing the singularity at \(\gamma=\gamma_0\). At the opposite limit,
\begin{equation}
\rho_0 T\xrightarrow[\rho_0\to\infty]{}\frac{c}{\alpha}.
\label{eq:T_large_rho_main}
\end{equation}
When the same \(\rho_0\) applies to the compared states, the induced ordering approaches that of the squared scale of the marginal Student--\(t\) law. Thus, \(\rho_0\) generates a continuous one-parameter interpolation between normalized displacement and marginal latent scale. Equation~\eqref{eq:rho0_fixed_marginal_main} shows that a fixed-marginal prior fiber explores only the bounded subrange \((0,2c_0)\) of this interpolation family.
\end{remark}

Since \(T>0\), \cref{eq:B_T_main} gives \(B>\nu_0/2\). Monotonicity therefore yields an analytic amplitude bound.

\begin{corollary}[Prior-determined upper bound]
\label{cor:bound}
For every admissible quotient state,
\begin{equation}
0<\frac1{\nu_{\can}}<M_0,
\label{eq:bound_main}
\end{equation}
where
\begin{equation}
M_0=
\frac{
\sqrt{(\nu_0-2\alpha_0-1)^2+4\nu_0}
-(\nu_0-2\alpha_0-1)
}{2\nu_0}.
\label{eq:M0_main}
\end{equation}
The bound depends only on \((\nu_0,\alpha_0)\), not on \((\gamma_0,\beta_0)\). For fixed \(\nu_0\), \(M_0\) is strictly increasing in \(\alpha_0\) and
\begin{equation}
\lim_{\alpha_0\downarrow0}M_0=\frac1{\nu_0}.
\label{eq:M0_limit_main}
\end{equation}
Hence, \(1/\nu_0\) is the infimum of the ceiling over \(\alpha_0>0\). At the opposite limit,
\begin{equation}
M_0
=\frac{2\alpha_0+1}{\nu_0}-1
+O(\alpha_0^{-1}),
\qquad \alpha_0\to\infty,
\label{eq:M0_large_alpha_main}
\end{equation}
so the ceiling is finite for every fixed prior but is not uniformly bounded as the prior tail parameter \(\alpha_0\) varies. Whenever \(\alpha>1\), the ratio \(\uepi/\uvar\) has the same prior-determined upper bound.
\end{corollary}

The reparameterization in \cref{eq:prior_reparam_main} makes the complementary roles explicit: \((\gamma_0,\rho_0)\) determine ordinal geometry, whereas \((\nu_0,\alpha_0)\) determine the hard ceiling \(M_0\). The complete prior determines the numerical transfer \cref{eq:transfer_main} between them.

For fixed \((\gamma,\alpha)\) and prior, \(c\downarrow0\) implies \(T\downarrow0\), \(\nu_{\can}\to\infty\), and \(\uepi/\uvar\to0\). Conversely, \(c\to\infty\) implies \(T\to\infty\) and \(1/\nu_{\can}\uparrow M_0\). Thus, increasing latent spread shifts the canonical allocation toward location uncertainty, but only up to a ceiling imposed by the prior.

When \(\alpha>1\), the same coordinate may be written using the actual latent variance:
\begin{equation}
T
=
\frac{\alpha-1}{\alpha}
\frac{\Var(z\mid\bm y)}
{(\gamma-\gamma_0)^2+\rho_0}.
\label{eq:T_var_main}
\end{equation}
For \(\alpha\leq1\), \cref{eq:T_main} remains a valid quotient coordinate but should not be interpreted through a finite second moment.

\begin{remark}[Latent-scale covariance and weight scope]
\label{rem:scale_scope}
The ordinal coordinate is tied to the scale convention set by the prior. Under a consistent change of latent units \(z\mapsto\kappa z\), with \(\gamma\mapsto\kappa\gamma\), \(\gamma_0\mapsto\kappa\gamma_0\), \(c\mapsto\kappa^2c\), and \(\rho_0\mapsto\kappa^2\rho_0\), the value of \(T\) is unchanged. Rescaling the quotient state while holding the prior fixed generally changes \(T\). Accordingly, the conditional weight invariance in Remark~\ref{rem:immediate} does not imply that ordering is invariant across outer optima obtained with different regularization weights.
\end{remark}

The ordinal equivalences above are not statements about arbitrary nonlinear aggregation across latent coordinates. Averaging, taking extrema, or applying other nonlinear summaries to coordinatewise scores can destroy rank equivalence and can reintroduce dependence on the numerical calibration set by \((\nu_0,\alpha_0)\).

\subsection{Likelihood blindness and variational suboptimality off the section}
Because every reconstruction functional of the marginal latent law is constant along a fiber, an observation-space residual cannot distinguish two representatives with the same quotient coordinates. Any additional criterion that acts directly on the fiber coordinate therefore supplies information not contained in the reconstruction likelihood.

Theorem~\ref{thm:reduction} gives a sharper variational statement. At a fixed quotient state, every noncanonical fiber representative has exactly the same reconstruction term but strictly larger forward KL to the chosen prior. Thus a noncanonical allocation is not merely likelihood-unidentified; it is variationally suboptimal for the hierarchical objective defined in \cref{eq:J4}. An auxiliary fiber criterion may define a different objective, but if it selects a noncanonical representative, that point necessarily has lower hierarchical ELBO than the canonical representative at the same quotient state.

\section{Discussion and Conclusion}
\label{sec:discussion}

The construction is most naturally viewed as a quotient-and-section problem. A four-parameter NIG hierarchy maps many-to-one onto a three-parameter Student--\(t\) latent law. The fibers are the curves \(\beta=c\nu/(1+\nu)\), along which every exact reconstruction functional is unchanged. The quotient therefore removes one exact likelihood redundancy. While the marginal degeneracy itself is familiar from observation-space evidential regression \citep{amini2020,meinert2023}, the present contribution is the explicit latent quotient, the prior-relative variational section, and the exact reduction obtained by partial minimization.

The quotient alone does not provide a preferred representative. A distinguished section appears only after the variational reference geometry is specified. The complete hierarchical prior contains a genuine residual choice of fiber position, whereas the forward KL is singled out by the fixed variational objective: minimizing that same divergence on each fiber is precisely what makes the reduction variationally exact. The level of regularization matters as well. The present construction regularizes the full NIG hierarchy; regularizing only the collapsed quotient coordinates would remove the fiber from the objective altogether and define a different variational model rather than a different section of the same one. Another divergence can likewise define another section, but a different section is off-optimum for the original forward-KL objective at fixed quotient coordinates.

The principal consequence is that the selected allocation variable does not provide a fourth learned information coordinate. Equation~\eqref{eq:transfer_main} shows explicitly that
\begin{equation}
\frac1{\nu_{\can}}=g_{\nu_0,\alpha_0}(T),
\end{equation}
where \(T\) is already determined by \((\gamma,\alpha,c)\) and the prior. Thus, on the variationally selected section, the commonly reported epistemic-to-conditional ratio \(\uepi/\uvar\) is a calibrated re-expression of quotient geometry and an indirectly learned evidential channel. For rank-only applications under a common calibration map, the hierarchy can be reduced even further operationally: the score \(T\) alone produces the same ordering, while \(\nu_{\can}\) remains relevant when numerical allocation, uncertainty decomposition, or prior-relative calibration is required.

The residual prior gauge is not neutralized by this ordinal result. Its effect is instead characterized exactly. Ordering depends on the prior through \((\gamma_0,\rho_0)\), with \(\rho_0=2\beta_0/\nu_0\), while numerical calibration and the inverse-allocation ceiling depend additionally on \((\nu_0,\alpha_0)\). If a prior is moved along its own fiber while its marginal Student--\(t\) law is held fixed, then
\begin{equation}
\rho_0=\frac{2c_0}{1+\nu_0}\in(0,2c_0),
\end{equation}
and the ordering can change. Hence, the input-dependent functional freedom of \(\nu(\bm y)\) is replaced by a global reference gauge whose ordinal effect is compressed to two prior combinations---or to the single positive scalar \(\rho_0\) under the centered convention \(\gamma_0=0\). The bound \(\rho_0<2c_0\) also shows that moving only along a fixed marginal-prior fiber can approach the inverse-normalized-radius endpoint of the interpolation family but cannot reach the pure marginal-scale endpoint.

This scalar has a direct geometric interpretation. It regularizes the normalized displacement \((\gamma-\gamma_0)^2\) in \(T\) and continuously moves the ordinal coordinate between the quotient normalized-radius score \(c/[\alpha(\gamma-\gamma_0)^2]\) and the marginal Student--\(t\) squared scale \(c/\alpha\). Separately, the amplitude ceiling of inverse allocation depends only on \((\nu_0,\alpha_0)\). In the equivalent prior coordinates \((\gamma_0,\rho_0,\nu_0,\alpha_0)\), ordinal geometry and numerical calibration are therefore cleanly separated.

Relative to the original \ELVAE formulation \citep{wang2026elvae}, the main change is conceptual. A stationary allocation used there as a diagnostic of a four-coordinate model becomes here the defining section of a three-coordinate variational family. The resulting partial minimization has the same infimum as the full NIG variational objective, while any off-section representative is strictly suboptimal at fixed quotient coordinates. While the earlier DER literature resolves the corresponding likelihood degeneracy by adding an auxiliary evidence signal, here the fiber is instead resolved by the KL term already contained in the hierarchical variational objective. These are different ways of supplying information that the marginal likelihood itself cannot provide.

Several boundaries should remain explicit. First, the no-loss theorem concerns the underlying variational family after exact partial minimization, but finite amortized neural networks can have different approximation and optimization behavior. Second, the reduction assumes that the decoder accesses the hierarchy only through the induced latent variable \(z\), since a decoder or auxiliary term that directly reads higher-order variables changes the quotient structure. Third, for a fixed latent coordinate, rank equivalence holds across states under its fixed prior, and cross-dimensional ranking additionally requires a shared calibration map across the compared dimensions. Fourth, nonlinear aggregation can alter coordinatewise ordinal relations. Fifth, the quotient construction requires only \(\alpha>0\), whereas second-moment language requires \(\alpha>1\); moreover, large \(\alpha\) can make the tail coordinate statistically weak even though it is not exactly redundant. Sixth, the ordinal coordinate is invariant under a consistent rescaling of posterior and prior latent units, not under rescaling of the posterior against a fixed prior. Consequently, conditional invariance of the selector to the KL weight does not imply invariance of outer trained solutions.

Within these assumptions, the fourth NIG coordinate need not be an independently learned input-dependent degree of freedom. The marginal law identifies a three-dimensional quotient so that the complete hierarchical prior and the forward KL term of the fixed variational objective select a unique analytic section. On that section, inverse allocation is an explicit monotone transfer of the quotient score \(T\), making its information content, calibration, and residual gauge transparent.

\paragraph{Author--AI Collaboration.}
Generative AI tools, ChatGPT and Claude, were used to assist the author in brainstorming, formulation, analysis, drafting, checking, and polishing. 
The author conceptualized the idea and frame, supervised the AI models, and takes responsibility for the content.

\clearpage
\appendix
\section{Gaussian Collapse, Student--\texorpdfstring{$t$}{t} Marginal, and Variance Decomposition}
\label{app:collapse}

Write the hierarchy with independent \(\epsilon_1,\epsilon_2\sim\Normal(0,1)\):
\begin{align}
\mu &= \gamma+\sqrt{\frac{\sigma^2}{\nu}}\,\epsilon_1,\label{eq:app_mu_reparam}\\
z &= \mu+\sqrt{\sigma^2}\,\epsilon_2.
\label{eq:app_z_reparam}
\end{align}
Substitution gives
\begin{equation}
z=\gamma+\sqrt{\sigma^2}
\left(\frac{\epsilon_1}{\sqrt\nu}+\epsilon_2\right).
\end{equation}
Since the bracketed variable is Gaussian with variance \(1+1/\nu\),
\begin{equation}
z\mid\sigma^2
\sim
\Normal\!\left(
\gamma,
\sigma^2\left(1+\frac1\nu\right)
\right).
\label{eq:app_collapsed_cond}
\end{equation}
Define
\begin{equation}
\omega^2=\sigma^2\left(1+\frac1\nu\right),
\qquad
c=\beta\left(1+\frac1\nu\right).
\end{equation}
The inverse-gamma scale property gives the collapsed hierarchy
\begin{equation}
\omega^2\sim\IG(\alpha,c),
\qquad
z\mid\omega^2\sim\Normal(\gamma,\omega^2),
\label{eq:app_collapsed_hierarchy}
\end{equation}
which is independent of the fiber coordinate. Hence the quotient equivalence admits a pathwise representation that is exactly fiber-invariant.

Integrating \(\omega^2\) yields
\begin{equation}
z\mid\bm y\sim t_{2\alpha}\!\left(\gamma,\frac{c}{\alpha}\right),
\qquad \alpha>0,
\end{equation}
under the squared-scale convention used in the main text. Thus \((\gamma,\alpha,c)\) completely determine the marginal latent law.

For the second-moment interpretation only, assume \(\alpha>1\). Then
\begin{equation}
\E[\sigma^2]=\frac{\beta}{\alpha-1},
\end{equation}
and the law of total variance gives
\begin{align}
\Var(z\mid\bm y)
&=\E[\Var(z\mid\mu,\sigma^2)]+\Var(\E[z\mid\mu,\sigma^2])\\
&=\E[\sigma^2]+\Var(\mu)\\
&=\frac{\beta}{\alpha-1}
+\frac{\beta}{\nu(\alpha-1)}\\
&=\frac{c}{\alpha-1}.
\end{align}
Along the fiber \(\beta=c\nu/(1+\nu)\),
\begin{align}
\uvar
&=\frac{c}{\alpha-1}\frac{\nu}{1+\nu},\\
\uepi
&=\frac{c}{\alpha-1}\frac{1}{1+\nu}.
\end{align}
Evaluation at \(\nu=\nu_{\can}\) gives \cref{eq:decomp_main}.

\section{NIG Divergence and Restriction to a Fiber}
\label{app:kl}

Use the inverse-gamma parameterization
\begin{equation}
p(x\mid\alpha,\beta)
=\frac{\beta^\alpha}{\Gamma(\alpha)}
 x^{-\alpha-1}\exp\!\left(-\frac{\beta}{x}\right),
\qquad x>0.
\label{eq:app_ig_density}
\end{equation}
Then
\begin{equation}
\E[\log x]=\log\beta-\psi(\alpha),
\qquad
\E[x^{-1}]=\frac{\alpha}{\beta}.
\end{equation}
For
\begin{equation}
q=\NIG(\gamma,\nu,\alpha,\beta),
\qquad
p_0=\NIG(\gamma_0,\nu_0,\alpha_0,\beta_0),
\end{equation}
the chain rule for KL divergence gives
\begin{align}
\KL(q\|p_0)
=&\;\KL\!\left[\IG(\alpha,\beta)\|\IG(\alpha_0,\beta_0)\right]\nonumber\\
&+\E_{q(\sigma^2)}\KL\!\left[
\Normal\!\left(\gamma,\frac{\sigma^2}{\nu}\right)
\bigg\|
\Normal\!\left(\gamma_0,\frac{\sigma^2}{\nu_0}\right)
\right].
\label{eq:app_kl_split}
\end{align}
Direct calculation yields
\begin{align}
\KL(q\|p_0)
=&\;\alpha_0\log\frac{\beta}{\beta_0}
-\log\Gamma(\alpha)+\log\Gamma(\alpha_0)
+(\alpha-\alpha_0)\psi(\alpha)-\alpha
+\frac{\alpha\beta_0}{\beta}\nonumber\\
&+\frac12\left[
\log\frac{\nu}{\nu_0}
+\frac{\nu_0}{\nu}-1
+\frac{\nu_0\alpha(\gamma-\gamma_0)^2}{\beta}
\right].
\label{eq:app_nig_kl}
\end{align}

Restrict now to the fiber
\begin{equation}
\beta=c\frac{\nu}{1+\nu}.
\end{equation}
Collecting all terms independent of \(\nu\) into a constant \(C\), the remaining one-dimensional function is
\begin{equation}
\mathcal R(\nu)
=
\left(\alpha_0+\frac12\right)\log\nu
-\alpha_0\log(1+\nu)
+\frac{B}{\nu}+C,
\label{eq:app_Rnu}
\end{equation}
where
\begin{equation}
B=
\frac{\nu_0}{2}
+\frac{\alpha}{c}
\left[
\beta_0+\frac{\nu_0}{2}(\gamma-\gamma_0)^2
\right].
\label{eq:app_B}
\end{equation}
For completeness, the logarithmic reduction is
\begin{align}
\alpha_0\log\beta+\frac12\log\nu
&=\alpha_0\log c
+\left(\alpha_0+\frac12\right)\log\nu
-\alpha_0\log(1+\nu),
\end{align}
and the reciprocal terms satisfy
\begin{align}
\frac{\alpha\beta_0}{\beta}
+\frac{\nu_0\alpha(\gamma-\gamma_0)^2}{2\beta}
+\frac{\nu_0}{2\nu}
=
C_1+\frac{B}{\nu},
\end{align}
for a \(\nu\)-independent constant \(C_1\). This establishes \cref{eq:app_Rnu}.

\subsection{Closed form of the reduced regularizer}
Let \(\nu_{\can}\) be the analytic root in \cref{eq:nucan_main} and
\begin{equation}
\beta_{\can}=c\frac{\nu_{\can}}{1+\nu_{\can}}.
\end{equation}
Then the partially minimized divergence in \cref{eq:Rcan_main} is explicitly
\begin{align}
\Rcan(\gamma,\alpha,c)
=&\;\alpha_0\log\frac{\beta_{\can}}{\beta_0}
-\log\Gamma(\alpha)+\log\Gamma(\alpha_0)
+(\alpha-\alpha_0)\psi(\alpha)-\alpha
+\frac{\alpha\beta_0}{\beta_{\can}}\nonumber\\
&+\frac12\left[
\log\frac{\nu_{\can}}{\nu_0}
+\frac{\nu_0}{\nu_{\can}}-1
+\frac{\nu_0\alpha(\gamma-\gamma_0)^2}{\beta_{\can}}
\right].
\label{eq:app_Rcan_closed}
\end{align}
Because \(\nu_{\can}\) and \(\beta_{\can}\) are explicit functions of \((\gamma,\alpha,c)\), \cref{eq:app_Rcan_closed} is a closed-form function of the quotient state and fixed prior alone.

\section{Proof of the Prior-Relative Selector and Its Geometry}
\label{app:canonical}

Differentiating \cref{eq:app_Rnu} gives
\begin{equation}
\mathcal R'(\nu)
=
\frac{\alpha_0+1/2}{\nu}
-\frac{\alpha_0}{1+\nu}
-\frac{B}{\nu^2}.
\label{eq:app_Rprime}
\end{equation}
Multiplication by \(2\nu^2(1+\nu)>0\) gives
\begin{equation}
\nu^2+(2\alpha_0+1-2B)\nu-2B=0.
\label{eq:app_quad}
\end{equation}
The roots have product \(-2B<0\), so exactly one is positive. The positive root is
\begin{equation}
\nu_{\can}
=B-\alpha_0-\frac12
+\sqrt{\left(B-\alpha_0-\frac12\right)^2+2B}.
\end{equation}
Moreover,
\begin{equation}
\mathcal R(\nu)\to\infty\quad(\nu\downarrow0),
\qquad
\mathcal R(\nu)\to\infty\quad(\nu\to\infty),
\end{equation}
so the unique positive stationary point is the unique global minimum.

\subsection{Real analyticity and derivative bounds}
Let \(a=\alpha_0+1/2\) and
\begin{equation}
D(B)=(B-a)^2+2B.
\end{equation}
Because \(B>0\), the discriminant satisfies \(D(B)>0\) immediately, so the positive square root is real analytic. Since \(B\) is real analytic for \(c>0\), \(\nu_{\can}\) is real analytic on \(\mathbb R\times(0,\infty)\times(0,\infty)\).

For the sharper derivative bounds, use the identity
\begin{align}
D(B)-(B-a+1)^2
&=2a-1\\
&=2\alpha_0>0.
\label{eq:app_identity}
\end{align}
Differentiating gives
\begin{equation}
\frac{d\nu_{\can}}{dB}
=1+\frac{B-a+1}{\sqrt{(B-a)^2+2B}}.
\label{eq:app_dnudB}
\end{equation}
By \cref{eq:app_identity}, the denominator strictly exceeds \(|B-a+1|\), hence
\begin{equation}
0<\frac{d\nu_{\can}}{dB}<2.
\label{eq:app_dnu_bounds}
\end{equation}

\subsection{Prior self-consistency and prior-fiber dependence}
Let
\begin{equation}
c_0=\beta_0\left(1+\frac1{\nu_0}\right).
\end{equation}
At \((\gamma,\alpha,c)=(\gamma_0,\alpha_0,c_0)\), the point \((\nu,\beta)=(\nu_0,\beta_0)\) belongs to the corresponding fiber and makes the divergence zero. Uniqueness therefore gives
\begin{equation}
\nu_{\can}=\nu_0,
\qquad
\beta_{\can}=\beta_0.
\end{equation}

Self-consistency does not make \(\nu_0\) intrinsic. Holding the prior marginal coordinates \((\gamma_0,\alpha_0,c_0)\) fixed gives
\begin{equation}
\beta_0=\frac{c_0\nu_0}{1+\nu_0},
\end{equation}
and therefore
\begin{equation}
B
=
\frac{\nu_0}{2}
\left[
1+\frac{\alpha}{c}
\left(
(\gamma-\gamma_0)^2+\frac{2c_0}{1+\nu_0}
\right)
\right].
\label{eq:app_prior_fiber_B}
\end{equation}
Equivalently, along this fixed-marginal prior fiber,
\begin{equation}
\rho_0=\frac{2\beta_0}{\nu_0}=\frac{2c_0}{1+\nu_0}.
\label{eq:app_rho_fixed_marginal}
\end{equation}
Since \(\nu_0\in(0,\infty)\),
\begin{equation}
0<\rho_0<2c_0,
\qquad
\rho_0\downarrow0\;(\nu_0\to\infty),
\qquad
\rho_0\uparrow2c_0\;(\nu_0\downarrow0).
\label{eq:app_rho_range}
\end{equation}
Thus moving the prior fiber position changes the ordinal parameter \(\rho_0\) as well as the numerical selector, even though the prior marginal Student--\(t\) law is unchanged. It can approach the \(\rho_0=0\) endpoint but cannot reach arbitrarily large \(\rho_0\) while preserving that marginal law.

\subsection{Geometric coordinate and transfer function}
Define
\begin{equation}
\rho_0=\frac{2\beta_0}{\nu_0},
\qquad
T
=
\frac{c}
{\alpha\left[(\gamma-\gamma_0)^2+\rho_0\right]}.
\end{equation}
Then
\begin{equation}
B
=\frac{\nu_0}{2}\left(1+\frac1T\right).
\end{equation}
The map \(T\mapsto B\) is strictly decreasing. By \cref{eq:app_dnu_bounds}, \(B\mapsto\nu_{\can}\) is strictly increasing and hence \(B\mapsto1/\nu_{\can}\) is strictly decreasing. Therefore the composition of the two strictly decreasing maps,
\begin{equation}
T\longmapsto B\longmapsto\frac1{\nu_{\can}(B)},
\end{equation}
is strictly increasing. This proves Proposition~\ref{prop:geometry}.

The displayed transfer function follows by rationalizing the reciprocal. With \(d=B-a\),
\begin{align}
\frac1{\nu_{\can}}
&=\frac1{d+\sqrt{d^2+2B}}\\
&=\frac{\sqrt{d^2+2B}-d}{2B},
\end{align}
which is \cref{eq:transfer_main} after substituting \(B=B(T)\). The two algebraic forms should be evaluated branchwise for numerical stability: \(d+\sqrt{d^2+2B}\) is stable for \(d\ge0\), whereas \(2B/[\sqrt{d^2+2B}-d]\) is stable for \(d<0\). Corollary~\ref{cor:noextra} then follows directly: on the selected section both \(1/\nu_{\can}\) and, for \(\alpha>1\), the variance-allocation ratios are deterministic functions of quotient coordinates and the fixed prior.

\subsection{Ordinal dependence on the prior}
The positive change of variables
\begin{equation}
(\gamma_0,\nu_0,\alpha_0,\beta_0)
\longleftrightarrow
(\gamma_0,\rho_0,\nu_0,\alpha_0),
\qquad
\beta_0=\frac{\rho_0\nu_0}{2},
\end{equation}
is bijective. For a fixed complete prior, ordering is the ordering by
\begin{equation}
T=\frac{c}{\alpha[(\gamma-\gamma_0)^2+\rho_0]}.
\end{equation}
Hence only \((\gamma_0,\rho_0)\) enter the ordinal coordinate. If two complete priors share these two quantities, they assign the same \(T\) to every quotient state. Their transfer functions from \(T\) to \(1/\nu_{\can}\) may differ through \((\nu_0,\alpha_0)\), but each transfer function is strictly increasing. They therefore induce exactly the same ordering over quotient states under each fixed prior. This proves Corollary~\ref{cor:ordinal}.

This statement is transverse to the fixed-marginal prior slice considered above. If \((\gamma_0,\alpha_0,c_0)\) is fixed and \(\nu_0\) is varied, then \cref{eq:app_rho_fixed_marginal} shows that \(\rho_0\) changes. Corollary~\ref{cor:ordinal} therefore gives no invariance along that slice; ranks may change.

For comparisons across different latent dimensions, an additional condition is needed. If dimensions \(k\) have different calibration pairs \((\nu_{0,k},\alpha_{0,k})\), then their maps \(g_k(T_k)\) differ and a cross-dimensional ordering by \(T_k\) need not agree with the ordering by \(1/\nu_{\can,k}\). A shared coordinatewise prior, or at least a common \((\nu_0,\alpha_0)\), makes the transfer function common and restores exact rank equivalence across dimensions.

\subsection{Interpolation induced by the prior radius floor}
Write
\begin{equation}
T_{\rho_0}
=\frac{c}{\alpha[(\gamma-\gamma_0)^2+\rho_0]}.
\end{equation}
For \(\gamma\neq\gamma_0\),
\begin{equation}
\lim_{\rho_0\downarrow0}T_{\rho_0}
=\frac{c}{\alpha(\gamma-\gamma_0)^2}.
\end{equation}
If \(\gamma=\gamma_0\), the unregularized endpoint diverges, so positive \(\rho_0\) acts as a radius floor. At the other endpoint,
\begin{equation}
\lim_{\rho_0\to\infty}\rho_0T_{\rho_0}
=\frac{c}{\alpha}.
\end{equation}
For a common \(\rho_0\) across compared states, multiplication by the positive scalar \(\rho_0\) does not alter ranking. Hence the large-\(\rho_0\) ordinal limit is the ranking by the marginal Student--\(t\) squared scale \(c/\alpha\). When \(\alpha>1\), the small-\(\rho_0\) endpoint can also be written
\begin{equation}
\frac{c}{\alpha(\gamma-\gamma_0)^2}
=
\frac{\alpha-1}{\alpha}
\frac{\Var(z\mid\bm y)}{(\gamma-\gamma_0)^2}.
\end{equation}
Equation~\eqref{eq:app_rho_range} shows that a fixed-marginal prior fiber explores only a bounded subrange of this interpolation family.

\subsection{Latent-scale covariance}
Under a consistent rescaling of latent units \(z\mapsto\kappa z\), the quotient and prior coordinates transform as
\begin{equation}
\gamma\mapsto\kappa\gamma,
\qquad
\gamma_0\mapsto\kappa\gamma_0,
\qquad
c\mapsto\kappa^2c,
\qquad
\rho_0\mapsto\kappa^2\rho_0.
\end{equation}
Therefore
\begin{equation}
T\mapsto
\frac{\kappa^2c}
{\alpha[\kappa^2(\gamma-\gamma_0)^2+\kappa^2\rho_0]}
=T.
\end{equation}
Thus \(T\) is invariant to a consistent change of latent units, but not to a rescaling of the quotient state against a fixed prior. The prior fixes the reference scale.

\subsection{Upper bound and limiting behavior}
Because \(T>0\),
\begin{equation}
B>\frac{\nu_0}{2}.
\end{equation}
Strict monotonicity implies
\begin{equation}
\nu_{\can}(B)>
\nu_{\min}:=
\left.\nu_{\can}(B)\right|_{B=\nu_0/2}.
\label{eq:app_numin_def}
\end{equation}
Writing \(x=(\nu_0-2\alpha_0-1)/2\),
\begin{equation}
\nu_{\min}=x+\sqrt{x^2+\nu_0},
\end{equation}
and rationalization gives
\begin{equation}
\frac1{\nu_{\min}}
=
\frac{
\sqrt{(\nu_0-2\alpha_0-1)^2+4\nu_0}
-(\nu_0-2\alpha_0-1)
}{2\nu_0}
=M_0.
\end{equation}
This proves the bound in Corollary~\ref{cor:bound}. The closed form contains only \((\nu_0,\alpha_0)\), showing explicitly that the amplitude ceiling is independent of \((\gamma_0,\beta_0)\).

For the dependence on \(\alpha_0\), let
\begin{equation}
u=\nu_0-2\alpha_0-1.
\end{equation}
Then
\begin{equation}
M_0=\frac{\sqrt{u^2+4\nu_0}-u}{2\nu_0},
\end{equation}
and
\begin{equation}
\frac{\partial M_0}{\partial\alpha_0}
=
\frac{1}{\nu_0}
\left(
1-\frac{u}{\sqrt{u^2+4\nu_0}}
\right)>0.
\end{equation}
At the boundary \(\alpha_0\downarrow0\),
\begin{align}
M_0
&\longrightarrow
\frac{\sqrt{(\nu_0-1)^2+4\nu_0}-(\nu_0-1)}{2\nu_0}\\
&=\frac{(\nu_0+1)-(\nu_0-1)}{2\nu_0}
=\frac1{\nu_0}.
\end{align}
Hence \(1/\nu_0\) is the infimum of the ceiling over \(\alpha_0>0\).

For the opposite limit, set \(L=2\alpha_0+1-\nu_0=-u\). As \(\alpha_0\to\infty\),
\begin{equation}
\sqrt{L^2+4\nu_0}
=L+\frac{2\nu_0}{L}+O(L^{-3}),
\end{equation}
so
\begin{equation}
M_0
=\frac{L}{\nu_0}+O(L^{-1})
=\frac{2\alpha_0+1}{\nu_0}-1+O(\alpha_0^{-1}),
\end{equation}
which proves \cref{eq:M0_large_alpha_main}. Thus the ceiling is finite for each fixed complete prior but has no bound uniform in \(\alpha_0\).

Finally, for fixed \((\gamma,\alpha)\), \(c\downarrow0\) implies \(T\downarrow0\), \(B\to\infty\), and \(\nu_{\can}\to\infty\). Likewise, \(c\to\infty\) implies \(T\to\infty\), \(B\downarrow\nu_0/2\), and \(1/\nu_{\can}\uparrow M_0\).

\section{Proofs for the Reduced Variational Objective}
\label{app:reduction}

\subsection{Proof of Theorem~\ref{thm:reduction}}
For fixed \((\gamma,\alpha,c)\), Proposition~\ref{prop:quotient} implies that \(\Lrec(\gamma,\alpha,c;\bm y)\) is constant along the fiber. Therefore
\begin{align}
&\inf_{\nu>0,\;\beta=c\nu/(1+\nu)}
\left\{
\Lrec(\gamma,\alpha,c;\bm y)
+\lambda\KL(q\|p_0)
\right\}\\
&\qquad=
\Lrec(\gamma,\alpha,c;\bm y)
+\lambda
\inf_{\nu>0,\;\beta=c\nu/(1+\nu)}
\KL(q\|p_0)\\
&\qquad=
\Lrec(\gamma,\alpha,c;\bm y)
+\lambda\Rcan(\gamma,\alpha,c).
\end{align}
Taking the infimum over the quotient coordinates gives
\begin{equation}
\inf_{\gamma,\alpha,c,\nu}\mathcal J_4
=
\inf_{\gamma,\alpha,c}\mathcal J_3.
\end{equation}
Since the fiber minimizer is unique, any noncanonical point on a fixed fiber has strictly larger KL and hence strictly larger objective for \(\lambda>0\).

This is an identity of the underlying variational families after partial minimization. It does not require, and does not imply, equality between arbitrary finite amortized parameterizations used to represent those families.

\subsection{Negative-ELBO specialization}
For completeness, suppose the generative hierarchy factors as
\begin{equation}
p_0(\mu,\sigma^2)\,p(z\mid\mu,\sigma^2)\,p_\theta(\bm y\mid z)
\end{equation}
and the variational family as
\begin{equation}
q(\mu,\sigma^2\mid\bm y)\,q(z\mid\mu,\sigma^2,\bm y).
\end{equation}
The negative ELBO can be written
\begin{align}
-\mathrm{ELBO}
=&\;\E_q[-\log p_\theta(\bm y\mid z)]
+\KL\!\left[q(\mu,\sigma^2\mid\bm y)\|p_0(\mu,\sigma^2)\right]\nonumber\\
&+\E_{q(\mu,\sigma^2\mid\bm y)}
\KL\!\left[
q(z\mid\mu,\sigma^2,\bm y)
\|p(z\mid\mu,\sigma^2)
\right].
\label{eq:app_negative_elbo}
\end{align}
Under \cref{eq:conditional_match_main}, the final conditional KL vanishes exactly. Thus with \(\lambda=1\), \(\mathcal J_4\) is the negative ELBO and Theorem~\ref{thm:reduction} is exact partial minimization of that variational objective.

\subsection{Two immediate consequences}
For fixed \((\gamma,\alpha,c)\),
\begin{equation}
\arg\min_\nu \lambda\mathcal R(\nu)
=
\arg\min_\nu \mathcal R(\nu),
\qquad \lambda>0,
\end{equation}
which gives the conditional weight invariance stated in Remark~\ref{rem:immediate}. The outer optimum of the quotient coordinates can still change with \(\lambda\).

For the envelope identity, let \(F(\theta,\nu)\) denote the NIG divergence after substituting \(\beta=c\nu/(1+\nu)\), where \(\theta=(\gamma,\alpha,c)\). Then
\begin{equation}
\Rcan(\theta)=F(\theta,\nu_{\can}(\theta)).
\end{equation}
By Theorem~\ref{thm:canonical}, \(\nu_{\can}(\theta)\) is an interior real-analytic minimizer throughout the admissible domain, and
\begin{equation}
\partial_\nu F(\theta,\nu_{\can}(\theta))=0.
\end{equation}
Therefore
\begin{align}
\nabla_\theta\Rcan(\theta)
&=\partial_\theta F(\theta,\nu_{\can}(\theta))
+\partial_\nu F(\theta,\nu_{\can}(\theta))
\nabla_\theta\nu_{\can}(\theta)\\
&=\partial_\theta F(\theta,\nu_{\can}(\theta)),
\end{align}
which is \cref{eq:envelope_main}.

\end{document}